\documentclass[11pt]{article} 
\usepackage{hyperref}
\usepackage[margin=1in]{geometry} 
\usepackage{amsmath,amsthm,amssymb} 
\usepackage[shortlabels]{enumitem} 
\usepackage{booktabs} 
\usepackage{graphicx}
\usepackage{orcidlink}

\title{\textbf{Investigating GNN Convergence on Large Randomly Generated Graphs with Realistic Node Feature Correlations}}
\author{\textbf{Mohammed Zain Ali Ahmed\orcidlink{0000-0003-4420-4075}}\\ \\Email: mzain@live.co.uk}
\date{18/02/2026}
\begin{document}
\maketitle

\begin{abstract}
There are a number of existing studies analysing the convergence behaviour of graph neural networks on large random graphs. Unfortunately, the majority of these studies do not model correlations between node features, which would naturally exist in a variety of real-life networks. Consequently, the derived limitations of GNNs, resulting from such convergence behaviour, is not truly reflective of the expressive power of GNNs when applied to realistic graphs. In this paper, we will introduce a novel method to generate random graphs that have correlated node features. The node features will be sampled in such a manner to ensure correlation between neighbouring nodes. As motivation for our choice of sampling scheme, we will appeal to properties exhibited by real-life graphs, particularly properties that are captured by the Barabási-Albert model. A theoretical analysis will strongly indicate that convergence can be avoided in some cases, which we will empirically validate on large random graphs generated using our novel method. The observed divergent behaviour provides evidence that GNNs may be more expressive than initial studies would suggest, especially on realistic graphs.
\end{abstract}

\section{Introduction}

The convergence behaviour of GNNs on large random graphs has been analysed in various ways: Keriven et al. use comparisons towards continuous analogues in the limit \cite{keriven1, keriven2}, whereas Adam-Day et al. consider the output distribution when applied to increasingly larger random graphs \cite{zeroone, almostsurely}. In the latter of their works, Adam-Day et al. are able to provide a uniform bound on the expressive power of GNNs (where the model parameterisation is the same for graphs of all sizes). As mentioned in \cite{nonunif}, non-uniform bounds can be problematic in practice, as we may need to train the model repeatedly for larger graphs. Unfortunately, the analyses by Adam-Day et al. sample randomised node features independently, ignoring correlations that are likely to exist in practice. To better understand uniform bounds that may arise on realistic datasets, we will introduce a scheme for sampling correlated node features. We will modify the Barabási-Albert (BA) model \cite{bamodel}, which uses two mechanisms that have been observed in real-life networks: the idea of growth (that new nodes can join the graph at any time) and the idea of preferential attachment (that new nodes prefer to attach to high-degree nodes) \cite{bainfo2}. The latter of these principles can be seen as an example of assortative mixing \cite{newman}, a concept that often arises within real networks \cite{assort1, assort2}. This can be defined as the tendency for nodes to connect to ``similar" nodes, where similarity could be defined in terms of node degrees or some other scalar value. Assortativity has already been studied in the context of GNNs \cite{homophily1, homophily2, homophily3}, but most of this research is regarding node classification and focuses on analysing existing graphs as opposed to constructing them. Concurrently with the graph's construction (in the BA model), we will also be sampling node features for new nodes. This method of sampling will ensure positive correlations between features of neighbouring nodes with stronger correlations (preferential) to high-degree nodes, in line with assortative mixing. Here, the Pearson correlation \cite{pearson} will be used: $\text{corr}(X,Y) = \frac{\text{Cov}(X,Y)}{\sqrt{\text{Var}(X) \cdot \text{Var}(Y)}}$. Our work will take strong inspiration from the existing study by Adam-Day et al. \cite{almostsurely}. We will use a similar setup and model architecture, applying probabilistic GNN classifiers to the randomly generated graphs (with correlated node features) to observe whether there is convergence in the model outputs for large graphs, with the aim of expanding on the results shown in \cite{almostsurely}.


\section{Method}
We use a similar setup to \cite{almostsurely}, to expand on those results. We generate 30 graphs of each size, which we use to calculate mean and standard deviation outputs, with sizes varying from 25 to 2000. The graphs use 32-dimensional node features. A GNN (consisting of a 3-layer equal-width GAT or GCN \cite{gat, gcn} followed by mean pooling) is applied, which uses a ReLU and a weighted sum aggregation as in \cite{almostsurely}. The resulting vector is fed into a single-layer MLP and softmax activation to produce a 3-class distribution output for graph classification. The GNN classifier is not trained/modified and we use the randomised weights set by PyTorch Geometric, since we are examining uniform expressive power. Apart from varying the GNN (from GAT to GCN), we also vary the BA model and correlation scheme. From \cite{bainfo1}, if a graph $G \sim \text{BA}(n,m)$ then its average degree is approximately $2m$ for large $n$. When increasing $n$, we examine two cases: when $m = \Theta(1)$, for which $\lvert E \rvert = \Theta(\lvert V \rvert)$ and $G$ is sparse and when $m = \Theta(n)$ so that $G$ is dense as $\lvert E \rvert = \Theta(\lvert V \rvert^2)$. Alongside this, we also test 3 correlation methods: no correlations, simple correlations and rescaled correlations (see below). The aim is to examine model outputs for large graphs across all choices of architecture, correlation and density, in order to observe which scenarios indicate convergence of output distributions. These results will then be compared to \cite{almostsurely}.

The $\text{BA}(n,m)$ model \cite{bamodel} has many variations, we use the following: start with a complete graph of $m$ nodes. In one iteration, we add a new node and sample without replacement $m$ existing nodes to be neighbours of this new node. The probability of selecting $v_i$ as the first neighbour is $\frac{k_i}{\sum_{j} k_j}$, where $k_i$ is the current degree of $v_i$. After choosing $m$ neighbours, the next iteration begins. These iterations continue until we have $n$ nodes in total. In this paper, we augment the iterations by generating multivariate normal (MVN) $d$-dimensional node features $x_j \sim \text{MVN}(\mathbf{0}_d, \mathbf{I}_d)$ for new nodes $v_j$ ($j = m+1, ..., n$) as they are introduced. The initial complete graph has $x_i \sim \text{MVN}(\mathbf{0}_d, \mathbf{I}_d)$ independently for $i=1, ..., m$. For simplicity, we assume below that $d=1$ (so that $x_j \sim \mathcal{N}(0,1)$) but in general if $d > 1$, we apply the below procedure independently to each dimension. Given a new node, say $v_{m+1}$, we first sample $m$ neighbours (which we call $v_1, ..., v_m$) in accordance with the BA model. Let $p_i$ be the probability of sampling node $v_i$ as the first neighbour. We associate a correlation $\rho_i := p_i$ with $v_i$ such that, for $i=1, ..., m$, we desire $\text{corr}(x_{m+1}, x_i) = \rho_i$ for the to-be-sampled $x_{m+1} \sim \mathcal{N}(0,1)$ -- this models positive correlation between neighbours (homophily), with stronger correlations towards high-degree nodes (preferential).

Let $\rho = (\rho_1, ..., \rho_m)^T$ be the row vector of desired correlations between each neighbouring feature and the new feature. We model the joint probability distribution $\mathbf{x} = (x_1, ..., x_m, x_{m+1}) \sim \text{MVN}(\mathbf{0}_{m+1}, \mathbf{\Sigma}_{m+1})$ where $\mathbf{\Sigma}_{m+1}$ is a symmetric, positive-definite covariance matrix. As $\text{Var}(x_i) = 1$, we have $\text{corr}(x_i, x_j) = \text{Cov}(x_i, x_j)$, so that we have the block form $\mathbf{\Sigma}_{m+1} = \left[\begin{array}{ c | c } \mathbf{I}_m & \rho \\ \hline \rho^T & 1 \end{array}\right]$. Here, we have simplified by assuming that $x_i, x_j$ are independent for $1 \leq i < j \leq m$. From \cite{mvn}, we see that the conditional distribution $x_{m+1} | (x_1, ..., x_m) \sim \mathcal{N}(\mathbf{\rho}^T (x_1, ..., x_m)^T, 1 - \mathbf{\rho}^T \mathbf{\rho})$, so all we need to do is sample according to this distribution, which satisfies the desired correlations. In $d$ dimensions, we can succinctly express feature vector $x_{m+1} | \mathbf{X} \sim \text{MVN}(\rho^T \mathbf{X}, (1 - \mathbf{\rho}^T \mathbf{\rho}) \mathbf{I}_d)$ for efficient computation, where $\mathbf{X}$ is an $m \times d$ node feature matrix. Note that in order for $\mathbf{\Sigma}_{m+1}$ to be positive-definite, we require that $\lvert \mathbf{\Sigma}_{m+1} \rvert = 1 - \mathbf{\rho}^T \mathbf{\rho} > 0$ which is ensured since elements of $\mathbf{\rho}$ are obtained from a distribution. This expression for $\lvert \mathbf{\Sigma}_{m+1} \rvert$ can be calculated by induction on $m$ in $\text{BA}(n,m)$: using Laplace expansion \cite{la}, $\lvert \mathbf{\Sigma}_{m+1} \rvert = \lvert \mathbf{\Sigma}_m \rvert - \rho_m^2$, which with base case $\lvert \mathbf{\Sigma}_1 \rvert$ = 1 leads to the above expression.

In practice, we may wish to sample from a bounded distribution such as the uniform distribution $\text{U}[0,1]$, like in \cite{almostsurely}. Fortunately, this is not too difficult to do: given $A \sim \mathcal{N}(0,1)$, note that $B = \Phi(A) \sim \text{U}[0,1]$ and vice-versa (where $\Phi$ is the CDF of $\mathcal{N}(0,1)$) since $\mathbb{P}(B < k) = \mathbb{P}(A < \Phi^{-1}(k)) = \Phi(\Phi^{-1}(k)) = k$, due to $\Phi$ being strictly increasing. The correlations are slightly affected by this transformation but from \cite{unifnormcorr}, we note that the relation between the two correlations is $u = \frac{6}{\pi} \sin^{-1}(\frac{n}{2})$ (where $u$ and $n$ are correlations for $\text{U}[0,1]$ and  $\mathcal{N}(0,1)$ respectively). So, by using $\Phi$ and this equation to transform from $\mathcal{N}(0,1)$ to $\text{U}[0,1]$ and back, we can sample from $\text{U}[0,1]$. For $d > 1$, we apply these transformations to each dimension separately. In this case, the initial graph will have features sampled from $\text{U}[0,1]^d$. We store features in $\text{U}[0,1]^d$, convert each of them to $\text{MVN}(\mathbf{0}, \mathbf{I}_d)$ when adding a new node, then convert the newly sampled feature from $\text{MVN}(\mathbf{0}_d, \mathbf{I}_d)$ to $\text{U}[0,1]^d$ and store it.

The above scheme will be referred to as ``simple" correlations. We will also rescale the correlations as part of the investigation: this means that, after initialising $\rho_i = p_i$, we set $\rho^* = \frac{\rho}{\sum_{i=1}^m \rho_i}$ so that the elements of $\rho^*$ add to 1. We investigate this since we expect $\left\lvert\lvert \rho \right\rvert\rvert \rightarrow 0$ as $n \rightarrow \infty$ if we don't rescale. The vector $\rho^*$ is used instead of $\rho$ for node feature sampling -- note that $\lvert \mathbf{\Sigma}_{m+1} \rvert > 0$ still holds.

\section{Results}

\subsection{Theoretical results}

We analyse the correlations generated in the late-stage BA iterations of our method (when the graph is almost fully constructed). As $n$ becomes large, we can regard more and more iterations as being late-stage, such that the below analysis becomes asymptotically valid for a non-negligible proportion of all iterations/nodes. We can then argue what results should arise in the limit as $n \rightarrow \infty$. 

Let $C_i = \rho_i$ be the correlation towards the $i$th chosen neighbour in a late-stage iteration. Note that $C_1$ may differ in distribution to $C_2, ..., C_m$ since we sample non-uniformly without replacement. Then, the probability of picking a degree $k$ node first is $\mathbb{P}\left(C_1 = \frac{k}{r}\right) = \frac{k n_k}{r}$ where $n_k$ is the number of existing nodes with degree $k$ and $r$ is the degree sum over all existing nodes (which can be written as a sum over degree sizes $r = \sum_j j n_j$). These quantities change over time as the graph is constructed, for late-stage iterations we can assume that the number of existing nodes $\sum_k n_k \approx n$. For $G \sim \text{BA}(n,m)$, there are some approximations that we can use when $n$ is large. As mentioned before, the average degree is approximately $2m$ such that $r \approx 2mn$. Furthermore, the degree distribution follows a power law \cite{bainfo2} so that $\frac{n_k}{n} \approx \frac{2m^2}{k^3}$ which implies $n_k \approx \frac{2m^2 n}{k^3} \approx \frac{mr}{k^3}$. Finally, we will approximate sums over a particular range by a corresponding integral over the same region -- this has justification \cite{sumint1, sumint2} when the summand is a decreasing non-negative function (which will always be true for us) and becomes a better approximation with wider regions.

We can approximate $\mathbb{E}(C_1) = \sum_{k=m}^n \left(\frac{k}{r} \times \frac{k n_k}{r}\right) \approx \sum_{k=m}^n \frac{k^2 mr}{r^2 k^3} = \sum_{k=m}^n \frac{m}{kr} \approx \sum_{k=m}^n \frac{1}{2kn}$. Estimating with an integral, this is $\frac{1}{2n} \int_{m}^n \frac{1}{k} dk = \frac{\log(n) - \log(m)}{2n} \rightarrow 0$ as $n \rightarrow \infty$ (regardless of whether $m = \Theta(1)$ or $\Theta(n)$). If the $\mathbb{E}(C_i)$ behave similarly to $\mathbb{E}(C_1)$ asymptotically (which we anticipate), the effect of using correlations vanishes for large $n$ in both dense and sparse graphs. This leaves us with the scenario of no correlations used, for which \cite{almostsurely} observed convergence.

We now consider rescaled correlations. Since the $C_i$ aren't necessarily identically distributed, we appeal to a more robust technique to estimate the mean rescaling factor $\mathbb{E}\left(\sum_i C_i\right)$. Consider this equivalent description of an iteration in our method: partition the existing nodes into groups, where we collect nodes of the same degree into the same group. Sample $m$ nodes without replacement according to the BA model and let $m_k$ be defined as the size of the sample that came from group $k$. The probability of picking the first node from group/degree $k$ is $\frac{k n_k}{r}$. With this setup, the sum of the $m$ correlations is $Q = \sum_i C_i = \sum_{k=m}^n \frac{k m_k}{r}$. The distribution of the $m_k$ turns out to be an instance of the multivariate Wallenius' noncentral hypergeometric distribution \cite{wallen3, wallen2, wallen1}, which samples $m$ items without replacement from a total of $n$ items (that are split into $c$ groups). The probability of picking from group $k$ first is $\frac{\omega_k n_k}{\sum_j \omega_j n_j}$ where $n_k$ is the size of group $k$ and $\omega_k$ is a weight associated with group $k$. If we set $\omega_k = k$ for all $k$, we obtain the above method. It is noted in \cite{wallen3} that we can approximate $\mu_k = \mathbb{E}(m_k)$ with the following procedure: find $\theta$ such that if we set $\mu_k = n_k (1 - e^{k\theta})$, we then have $0 \leq \mu_k \leq n_k$ for all $k$ (equivalently, $\theta < 0$) and $\sum_k \mu_k = m$. With such a $\theta$, the true value of $\mu_k$ is approximately given by $n_k (1 - e^{k\theta})$.

Clearly, $\theta = \theta(n)$ will be a function of $n$. Inspired by experimentation, we will try $\theta(n) = \log(1 - \frac{1}{2n})$. Then, $\sum \mu_k = \sum \left(n_k (1 - e^{k\theta})\right) \approx n - \sum n_k e^{k\theta}$. If this equals $m$, we need $\sum n_k e^{k\theta} \approx n - m$. We can approximate $e^{k\theta} = (1 - \frac{1}{2n})^k \approx 1 - \frac{k}{2n}$ by using the binomial series \cite{binomseries} (to which it converges as $\left\lvert \frac{1}{2n} \right\rvert < 1$), the error will be $\Theta\left(\left(\frac{1}{2n}\right)^2\right)$ which will vanish rapidly for large $n$. Then, $\sum n_k e^{k\theta} \approx mr \int_m^n \left(\frac{1}{k^3} - \frac{1}{2nk^2}\right) dk \approx 2m^2 n \left(\frac{1}{2m^2} - \frac{1}{2n^2} - \frac{1}{2mn} + \frac{1}{2n^2}\right) = n - m$ as desired. Using this $\theta$, $\mathbb{E}(Q) = \sum_{k=m}^n \frac{k \mu_k}{r} \approx \frac{1}{r} \sum_{k=m}^n \left(k n_k - k n_k e^{k\theta}\right) \approx 1 - m \cdot \sum_{k=m}^n \frac{e^{k\theta}}{k^2}$. The sum $\sum_k \frac{e^{k\theta}}{k^2} \approx \int_m^n \left(\frac{1}{k^2} - \frac{1}{2nk}\right) dk \approx \frac{1}{m} - \frac{1}{n} - \frac{\log\left(\frac{n}{m}\right)}{2n}$. So, we estimate $\mathbb{E}(Q) \approx \frac{m\left(2+\log\left(\frac{n}{m}\right)\right)}{2n} = \frac{2 + \log(u)}{2u}$ where $u = \frac{n}{m}$.

For sparse graphs, $u \rightarrow \infty$ and we expect that the mean of the rescaled correlations is close to $\frac{\mathbb{E}(C_1)}{\mathbb{E}(Q)} \approx \frac{1}{m}$ which is independent of $n$ (even if $n$ is large). We use weighted sum aggregations in the GNNs but $\text{Var}(x_1 + ... + x_m) = \sum_i \text{Var}(x_i) + \sum_{i \neq j} \text{Cov}(x_i, x_j)$ where the last term will be non-zero and independent of $n$. This indicates the possibility of high variance in the aggregations/messages so that divergence may be possible here. If we have a dense graph, $u = \frac{n}{m}$ is constant. Since $\mathbb{E}(Q)$ is a function of $u$ alone whilst $\mathbb{E}(C_1)$ is not, we expect each rescaled correlation (with mean $\approx \frac{1}{m}$) to still converge to $0$ as $m = \Theta(n)$. So, the effect of using correlations vanishes for large $n$ and we again have the scenario of no correlations used, so we anticipate convergence.

\subsection{Empirical results}

\begin{figure}[h!]
\centering
\includegraphics[width=16cm]{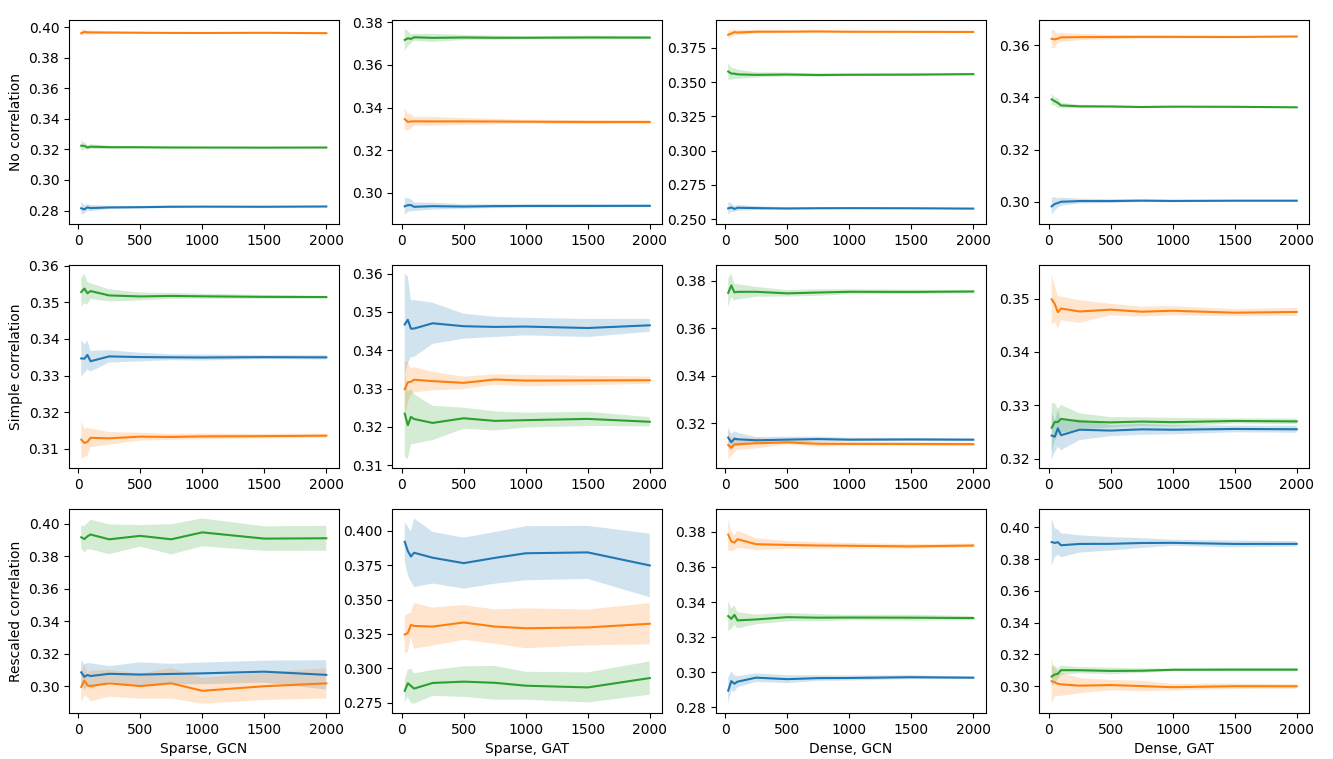}
\end{figure}

In the above figure, we plot graph sizes ($x$ axis) against classifier outputs ($y$ axis), for each of the 12 cases mentioned in the Methods. Each plot has 3 coloured lines, representing the mean probabilities for each class (averaged over 30 samples), with standard deviation shown in lower opacity. The rows correspond to the choice of correlation used. The columns correspond to different combinations of model (GAT or GCN) and graph density (sparse $G \sim BA(n,5)$ or dense $G \sim BA\left(n,\left\lfloor \frac{n}{5} \right\rfloor\right)$).

The empirical results are consistent with those of Adam-Day et al. \cite{almostsurely}. We observe convergence in all cases with no correlation used. As noted in \cite{almostsurely}, we observe that GAT exhibits higher variances than GCN (possibly due to the use of attention in aggregation). These results also support our theoretical findings: in sparse graphs with rescaled correlations (for GAT and GCN), no clear convergences are observed and we see constant high variance independent of $n$, which matches $\frac{\mathbb{E}(C_1)}{\mathbb{E}(Q)}$ also being independent of $n$. With simple correlations or dense graphs, convergence is observed (as expected).

The code used to produce these plots is available at \url{https://github.com/Zain123789/GNN-Project/blob/main/Project\%20Code.ipynb}.

\section{Conclusions}

Our findings provide evidence to support the claim that realistic correlations have the potential to affect GNN convergence (and also uniform expressiveness). Consequently, assuming independence of node features was crucial for the results obtained by Adam-Day et al. \cite{almostsurely} which suggested uniformly expressible properties must be asymptotically constant. To the contrary, our results indicate that, on realistic datasets with correlations, GNNs may be able to uniformly express properties that are not asymptotically constant, such as whether an undirected graph is acyclic (this implies $\lvert E \rvert \leq \lvert V \rvert - 1$) \cite{trees}. The probability of this holding on a random graph would likely approach 0 for large dense graphs, but not necessarily for large sparse graphs.

Despite the positive results, there are many limitations within our work, which could be the focus of further studies. One could investigate the effects of GNN components (such as non-linearity, aggregation, skip connections and width) on the final results. Our correlation scheme is specifically designed for continuous features, the use of one-hot (discrete) features would need a different scheme entirely. Using an autoregressive random graph generation model such as the BA model, we were able to simultaneously sample correlated features. But there are some limitations: for example, the BA model is known to have an unrealistic clustering coefficient \cite{bainfo2}, whereas the Watts-Strogatz model (based on rewiring edges) \cite{wsmodel} does not suffer from the same issue. Since this is not autoregressive in the same style, our scheme would need to be adapted here as well. We could also test our scheme on other autoregressive models, such as the Bianconi-Barábasi model \cite{bbmodel}. This augments the BA model with the concept of fitness (observed in real-life) which represents a natural ability to attract neighbours. Alternatively, autoregressive generative models such as GraphRNN and GRAN \cite{graphrnn, gran} attempt to learn a probability distribution over graphs but do not attempt to learn about the node features. Our scheme could be modified to support attempts for learning the joint adjacency and node feature matrix distribution. And finally, although many real-life networks are assortative, there are also many that are disassortative (heterophilic) \cite{assort1} so that dissimilar nodes tend to connect. We could modify our scheme by using negative correlations to investigate this model and we could also investigate making correlations anti-preferential (weaker correlations to high-degree nodes).

The analysis in our work is also imperfect. We make many approximations in the theoretical results, which need to be more carefully justified for a tight asymptotic analysis. In the Methods, we consider feature dimensions independently, which can be unrealistic for some applications. Furthermore, we also assume neighbouring features $x_i$ and $x_j$ are independent (for $1 \leq i < j \leq m$) when sampling $x_{m+1}$ but this is not true in practice. To be more precise, despite assuming independence when sampling, such feature dependencies would still exist in our model. This arises due to the BA model generating graphs with slowly growing average path lengths \cite{avgpath}, making it likely that feature $x_i$ could have been generated during a previous iteration involving feature $x_j$ (leading to a correlation between the two). Whilst our results show potential for improved uniform expressiveness, realistic datasets can exhibit a diverse range of properties and there are many future studies that need to be conducted to consolidate this research.

\bibliographystyle{unsrt}
\bibliography{GRLRefs.bib}

@inproceedings{almostsurely,
  title={Almost Surely Asymptotically Constant Graph Neural Networks},
  author={Adam-Day, Sam and Benedikt, Michael and Ceylan, Ismail Ilkan and Finkelshtein, Ben},
  booktitle={The Thirty-eighth Annual Conference on Neural Information Processing Systems},
  year={2024}
}

@article{gat,
  title={Graph attention networks},
  author={Veli{\v{c}}kovi{\'c}, Petar and Cucurull, Guillem and Casanova, Arantxa and Romero, Adriana and Lio, Pietro and Bengio, Yoshua},
  journal={arXiv preprint arXiv:1710.10903},
  year={2017}
}

@article{gcn,
  title={Semi-supervised classification with graph convolutional networks},
  author={Kipf, Thomas N and Welling, Max},
  journal={arXiv preprint arXiv:1609.02907},
  year={2016}
}

@article{bamodel,
  title={Emergence of scaling in random networks},
  author={Barab{\'a}si, Albert-L{\'a}szl{\'o} and Albert, R{\'e}ka},
  journal={science},
  volume={286},
  number={5439},
  pages={509--512},
  year={1999},
  publisher={American Association for the Advancement of Science}
}

@book{bainfo1,
author = {Barabási, Albert-László and Pósfai, Márton},
address = {Cambridge},
publisher = {Cambridge University Press},
title = {Network science},
year = {2016}
}

@article{bainfo2,
  title={Statistical mechanics of complex networks},
  author={Albert, R{\'e}ka and Barab{\'a}si, Albert-L{\'a}szl{\'o}},
  journal={Reviews of modern physics},
  volume={74},
  number={1},
  pages={47},
  year={2002},
  publisher={APS}
}

@book{mvn,
author = {Anderson, T. W. (Theodore Wilbur)},
address = {New York},
publisher = {Wiley},
series = {Wiley series in probability and mathematical statistics. Probability and mathematical statistics},
title = {An introduction to multivariate statistical analysis},
year = {1958}
}

@book{la,
  title={A guide to advanced linear algebra},
  author={Weintraub, Steven H},
  number={44},
  year={2011},
  publisher={MAA}
}

@book{unifnormcorr,
  title={On further methods of determining correlation},
  author={Pearson, Karl},
  volume={16},
  year={1907},
  publisher={Dulau and Company}
}

@article{sumint1,
  title={Summation Formulas of Euler--Maclaurin and Abel--Plana: Old and New Results and Applications},
  author={Milovanovi{\'c}, Gradimir V},
  journal={Progress in Approximation Theory and Applicable Complex Analysis: In Memory of QI Rahman},
  pages={429--461},
  year={2017},
  publisher={Springer}
}

@article{sumint2,
  title={A note on a generalization of the Cauchy-Maclaurin integral test},
  author={Newton, TA},
  journal={The American Mathematical Monthly},
  volume={61},
  number={5},
  pages={331--334},
  year={1954},
  publisher={JSTOR}
}

@book{wallen1,
  title={Biased sampling: the noncentral hypergeometric probability distribution.},
  author={Wallenius, Kenneth Ted},
  year={1964},
  publisher={Stanford University}
}

@article{wallen2,
  title={A non-central multivariate hypergeometric distribution arising from biased sampling with application to selective predation},
  author={Chesson, Jean},
  journal={Journal of Applied Probability},
  volume={13},
  number={4},
  pages={795--797},
  year={1976},
  publisher={Cambridge University Press}
}

@article{wallen3,
  title={Calculation methods for Wallenius' noncentral hypergeometric distribution},
  author={Fog, Agner},
  journal={Communications in Statistics—Simulation and Computation{\textregistered}},
  volume={37},
  number={2},
  pages={258--273},
  year={2008},
  publisher={Taylor \& Francis}
}

@book{binomseries,
author = {Geveci, Tunc.},
publisher = {Momentum Press},
title = {Intermediate calculus : infinite series},
year = {2016}
}

@article{zeroone,
  title={Zero-one laws of graph neural networks},
  author={Adam-Day, Sam and Ceylan, Ismail},
  journal={Advances in Neural Information Processing Systems},
  volume={36},
  pages={70733--70756},
  year={2023}
}

@article{keriven1,
  title={Convergence and stability of graph convolutional networks on large random graphs},
  author={Keriven, Nicolas and Bietti, Alberto and Vaiter, Samuel},
  journal={Advances in Neural Information Processing Systems},
  volume={33},
  pages={21512--21523},
  year={2020}
}

@article{keriven2,
  title={On the universality of graph neural networks on large random graphs},
  author={Keriven, Nicolas and Bietti, Alberto and Vaiter, Samuel},
  journal={Advances in Neural Information Processing Systems},
  volume={34},
  pages={6960--6971},
  year={2021}
}

@article{nonunif,
  title={Some might say all you need is sum},
  author={Rosenbluth, Eran and Toenshoff, Jan and Grohe, Martin},
  journal={arXiv preprint arXiv:2302.11603},
  year={2023}
}

@article{newman,
  title={Mixing patterns in networks},
  author={Newman, Mark EJ},
  journal={Physical review E},
  volume={67},
  number={2},
  pages={026126},
  year={2003},
  publisher={APS}
}

@article{assort1,
  title={Assortative mixing in networks},
  author={Newman, Mark EJ},
  journal={Physical review letters},
  volume={89},
  number={20},
  pages={208701},
  year={2002},
  publisher={APS}
}

@article{assort2,
  title={Birds of a feather: Homophily in social networks},
  author={McPherson, Miller and Smith-Lovin, Lynn and Cook, James M},
  journal={Annual review of sociology},
  volume={27},
  number={1},
  pages={415--444},
  year={2001},
  publisher={Annual Reviews 4139 El Camino Way, PO Box 10139, Palo Alto, CA 94303-0139, USA}
}

@article{pearson,
  title={Pearson correlation coefficient},
  author={Cohen, Israel and Huang, Yiteng and Chen, Jingdong and Benesty, Jacob and Benesty, Jacob and Chen, Jingdong and Huang, Yiteng and Cohen, Israel},
  journal={Noise reduction in speech processing},
  pages={1--4},
  year={2009},
  publisher={Springer}
}

@article{homophily1,
  title={Beyond homophily in graph neural networks: Current limitations and effective designs},
  author={Zhu, Jiong and Yan, Yujun and Zhao, Lingxiao and Heimann, Mark and Akoglu, Leman and Koutra, Danai},
  journal={Advances in neural information processing systems},
  volume={33},
  pages={7793--7804},
  year={2020}
}

@article{homophily2,
  title={Geom-gcn: Geometric graph convolutional networks},
  author={Pei, Hongbin and Wei, Bingzhe and Chang, Kevin Chen-Chuan and Lei, Yu and Yang, Bo},
  journal={arXiv preprint arXiv:2002.05287},
  year={2020}
}

@article{homophily3,
  title={Graph neural networks for graphs with heterophily: A survey},
  author={Zheng, Xin and Wang, Yi and Liu, Yixin and Li, Ming and Zhang, Miao and Jin, Di and Yu, Philip S and Pan, Shirui},
  journal={arXiv preprint arXiv:2202.07082},
  year={2022}
}

@book{trees,
  title={Discrete Mathematics with Graph Theory},
  author={Yadav, Santosh Kumar},
  year={2023},
  publisher={Springer}
}

@inproceedings{graphrnn,
  title={Graphrnn: Generating realistic graphs with deep auto-regressive models},
  author={You, Jiaxuan and Ying, Rex and Ren, Xiang and Hamilton, William and Leskovec, Jure},
  booktitle={International conference on machine learning},
  pages={5708--5717},
  year={2018},
  organization={PMLR}
}

@article{gran,
  title={Efficient graph generation with graph recurrent attention networks},
  author={Liao, Renjie and Li, Yujia and Song, Yang and Wang, Shenlong and Hamilton, Will and Duvenaud, David K and Urtasun, Raquel and Zemel, Richard},
  journal={Advances in neural information processing systems},
  volume={32},
  year={2019}
}

@article{wsmodel,
  title={Collective dynamics of ‘small-world’networks},
  author={Watts, Duncan J and Strogatz, Steven H},
  journal={nature},
  volume={393},
  number={6684},
  pages={440--442},
  year={1998},
  publisher={Nature Publishing Group}
}

@article{bbmodel,
  title={Competition and multiscaling in evolving networks},
  author={Bianconi, Ginestra and Barab{\'a}si, A-L},
  journal={Europhysics letters},
  volume={54},
  number={4},
  pages={436},
  year={2001},
  publisher={IOP Publishing}
}

@article{avgpath,
  title={The average path length of scale free networks},
  author={Chen, Fei and Chen, Zengqiang and Wang, Xiufeng and Yuan, Zhuzhi},
  journal={Communications in Nonlinear Science and numerical simulation},
  volume={13},
  number={7},
  pages={1405--1410},
  year={2008},
  publisher={Elsevier}
}

\end{document}